%% file: main.tex
\documentclass{article} 
\usepackage[final]{colm2026_conference}

\usepackage{microtype}
\usepackage{hyperref}
\usepackage{booktabs}
\usepackage{array}
\usepackage{amsmath}
\usepackage{amssymb}
\usepackage{colortbl}
\usepackage{graphicx}
\usepackage{subcaption}
\usepackage{stfloats}
\usepackage{placeins}
\usepackage{url}
\graphicspath{{figures/}}

\hypersetup{
  pdftitle={The World Won't Wait: Evaluating and Teaching Agents Under a Live Clock},
  pdfauthor={Ningxin Pan, Hanyu Li, Yeqing Shen, Ning Ding, Yehui Tang, Wentao Zhang}
}

\title{The World Won't Wait: Evaluating and Teaching Agents Under a Live Clock}

\author{
\mbox{Ningxin Pan$^{1,2}$, Hanyu Li$^{1}$, Yeqing Shen$^{2}$, Ning Ding$^{2}$, Yehui Tang$^{2,*}$,}\\
and Wentao Zhang$^{1,*}$\\
$^1$ Peking University \quad\quad $^2$ Samsung Research, Beijing, China\\
$^{*}$ Corresponding authors: \texttt{wentao.zhang@pku.edu.cn}, \texttt{yehui.tang@samsung.com}
}
\begin{document}
\maketitle

\begin{abstract}
Large language model (LLM) agents can solve complex interactive tasks, yet reasoning and tool calls take time. In continuously changing environments, an otherwise correct action may arrive too late. We introduce RealtimeBench, with 180 tasks across six game types, to evaluate agents while games continue running during inference. Paired evaluations show that GPT-6.1-sol outperforms the faster Jev when games are paused, but falls behind when they keep running. We propose Two-Timescale Offline Recursive Self-Improvement (TORSI): a slow teacher studies failures and tests corrections in paused games between attempts, then writes reusable instructions for a fast student. The student acts without teacher calls during gameplay, and validation scores determine which updates are adopted. Two rounds raise Jev's average test score from 13.51 to 24.19 on a 0--100 progress scale, improving all six game types. The instructions also improve average scores for three other models without retuning. These findings point toward real-time intelligence: enabling powerful LLM agents to remain effective in a continuously changing world by organizing deliberation, learning, and action across timescales.
\end{abstract}

\input{figures/figure-1}

\section{Introduction}

Large language model (LLM) agents can solve complex interactive tasks by combining reasoning with tool use. Bringing these capabilities into continuously changing environments presents a fundamental challenge: reasoning and tool calls take time, and the world can change before an agent finishes deciding what to do. An action that was appropriate when the agent observed the environment may arrive too late to be useful.

Real-time games provide controlled simulations of such dynamic environments, where agents must plan and act while the world keeps changing. Playing them well tests a core requirement for real-time intelligence: making good decisions and acting on them in time. For example, Mario may continue moving toward an enemy while the model decides whether to jump. If the command arrives after the collision, even a sensible decision cannot prevent failure (Figure~\ref{fig:realtime-example}).

However, game evaluations that pause during model responses remove these consequences of decision latency, so their scores may not predict performance when the game keeps moving. Prior work studies latency-sensitive decisions and games that continue during model computation \citep{kang2025win,zhang2025videogamebench,wen2026realtime}. Building on these efforts, we investigate how continued evolution during inference affects agent performance across different game mechanics. Comparing the same agent and tasks under different clocks allows us to examine when good action selection translates into effective real-time control.

To study this relationship, we introduce RealtimeBench, 180 tasks across six game types, including Mario, Snake, and Tetris. Models receive text descriptions of the current state and select from legal actions. While a response is pending, the game keeps running with the previous control still active. We also evaluate the same tasks with the game paused during each response or running faster. Scores measure in-game progress on a 0--100 scale, distinguishing partial achievement even when agents cannot complete a task or the game has no discrete completion condition.

These clock comparisons reveal how timing changes model performance. GPT-6.1-sol, which takes about 19 seconds per response, outperforms Jev, a faster decision model, in five of six game types when games are paused. When games keep running, their average scores are 1.51 and 13.51, respectively. Yet faster responses alone are insufficient: no evaluated model averages more than 16.62 (Table~\ref{tab:model-heldout}). We therefore investigate how a slower model's reasoning can improve a fast model's decisions without requiring the agent to wait for the slow model during gameplay.

To this end, we propose Two-Timescale Offline Recursive Self-Improvement (TORSI), which places action selection and learning on different timescales. A fast student chooses actions during gameplay, and a slow teacher studies its failures between attempts. The teacher tests corrections in paused environments and writes reusable instructions that the student follows in later games, without further teacher calls (Figure~\ref{fig:loops}). Instructions replace the current version only when they improve scores on separate validation tasks. The next round then addresses the updated student's remaining failures.

After two teaching rounds, the adopted instructions raise Jev's average test score from 13.51 to 24.19, improving all six game types. The same instructions improve average scores for three other models without retuning, although some game-specific regressions remain. Together, RealtimeBench and TORSI investigate how deliberation, learning, and action can be organized to support real-time intelligence. The benchmark measures what happens when the world continues to change during reasoning; the method explores how slow deliberation between attempts can improve fast action within it.

Our contributions are:
\begin{list}{\labelitemi}{%
  \setlength{\topsep}{4pt}%
  \setlength{\partopsep}{0pt}%
  \setlength{\itemsep}{2pt}%
  \setlength{\parsep}{0pt}%
}
\item \textbf{RealtimeBench:} We introduce a shared evaluation framework with 180 tasks across six game types and matched paused, live, and accelerated clocks. Explicit clock controls support consistent comparisons of response speed and performance under real-time conditions.
\item \textbf{Strong reasoning versus fast response:} We highlight the tension between reasoning capability and timely action as a central practical challenge for real-time agents, motivating the separation of deliberation and execution across timescales.
\item \textbf{TORSI:} We propose a method that iteratively turns offline LLM reasoning and practice into reusable guidance for a fast actor, with validation-gated updates and no teacher calls during gameplay.
\end{list}

\section{Related Work}

\paragraph{Game benchmarks and time.} SmartPlay, BALROG, and lmgame-Bench evaluate agents through step-based interaction \citep{wu2024smartplay,paglieri2025balrog,hu2026lmgame}. VideoGameBench includes both live and paused evaluation \citep{zhang2025videogamebench}, while Real-Time Reasoning Gym studies agents in continuously evolving environments \citep{wen2026realtime}. RealtimeBench explicitly controls the clock to support consistent comparisons between methods. This makes it easier to analyze how response speed affects task performance and whether a method remains effective under real-time conditions (Appendix~\ref{app:evidence}, Table~\ref{tab:benchmark-comparison}).

\paragraph{Acting while the world moves.} Real-time and delayed reinforcement learning study evolution during action selection \citep{ramstedt2019realtime,derman2021delayed}. For large models, approaches include latency--quality trade-offs and time-budget adaptation \citep{kang2025win,ma2026timely}, asynchronous inference \citep{riemer2024realtime}, and overlapping inference with action execution \citep{black2025rtc}. Hybrid agents combine fast control with slower planning during deployment \citep{lin2023swiftsage,christakopoulou2024talker,wen2026realtime,nvidia2025groot,shi2025hirobot}.

In games, GameSense uses VLMs to develop execution modules for real-time games \citep{lu2025gamesense}. TORSI keeps actor weights and interfaces fixed, revising guidance through offline practice and validation. Decision models such as Jev, Clef, and StartLux score predefined choices \citep{almeida2026jev,deusser2026jev,chen2026clef,startlux2026decision}, making them candidates for the fast actor.

\paragraph{Improvement without weight updates.} Agents learn through reflection, distilled insights, and skill libraries \citep{shinn2023reflexion,zhao2024expel,wang2023voyager}. GEPA evolves prompts from execution feedback, ACE refines reusable context, and AHE revises agent components \citep{agrawal2025gepa,zhang2025ace,lin2026ahe}. Recuris and RSIAgent explicitly iterate improvement through validated memory updates and verified environment knowledge \citep{yu2026recuris,zhu2026rsiagent}. TORSI adds offline practice for live-clock control, iterating guidance updates from the current student's failures.

\section{RealtimeBench}

RealtimeBench evaluates how well agents can act in games that continue running while they decide what to do.

\subsection{Live-Clock Protocol}

RealtimeBench keeps the game running while the agent selects an action from the current observation. The previous accepted control remains active during inference, so the observed state may be outdated when the new action takes effect.

We distinguish response latency from action age. For request $i$, let $t_i^o$ be the observation time, $t_i^r$ the response time, and $t_i^a$ the action-application time. Response latency measures the wait for a response; action age also includes any delay before the selected action is applied:
\begin{equation}
 L_i=t_i^r-t_i^o,\qquad A_i=t_i^a-t_i^o.
\end{equation}
The two differ when a response must wait for an eligible game tick or scheduled action. Outcomes follow the state at $t_i^a$ (Figure~\ref{fig:realtime-example}(b)).

Each game type declares its own wall-clock pacing; real-time evaluation therefore means continued evolution during inference, rather than a common frame rate across games. Paused and accelerated conditions change pacing without changing the underlying game physics.

\subsection{Game Types and Tasks}

The benchmark spans six game types with different control demands: platforming (Mario), navigation and food collection (Snake), falling-piece placement (Tetris), obstacle avoidance (Temple), steering and shooting (River), and timing flap actions (Bird; Figure~\ref{fig:overview}).

Within each game type, a \emph{task} fixes the game version, seed or stage, difficulty, action interface, and budget; an \emph{episode} is one execution of that task, ending at game termination or the budget. Settings and clocks for each game type are listed in Table~\ref{tab:suite}.

For development and evaluation, each game type has 30 tasks: ten feedback, ten validation, and ten test tasks, giving 180 tasks in total. Feedback tasks provide experience for developing an agent; validation tasks compare candidate updates; test tasks are held out from adaptation and used exclusively for final reporting.

Across these tasks, agents receive structured text descriptions rather than raw pixels. The \emph{harness}, the code connecting the model and game, constructs observations and legal action choices, queries the model, and applies its response (Figure~\ref{fig:base-harness}). It may compute geometry and timing features from the current state, but cannot inspect future states or search the simulator during scored play. Table~\ref{tab:observation-action-interfaces} in Appendix~\ref{app:evidence} summarizes the observations and action interfaces for each game type.

\input{figures/figure-2}
\input{figures/figure-base-harness}

\subsection{Scoring}
\label{sec:scoring}

We score these tasks by progress rather than binary completion for two reasons. Some games have no single completion condition and are naturally scored by points or survival progress. Even when completion is defined, a binary score cannot distinguish partial progress when successful completion is rare. Progress scores let us compare agents that improve without reaching a terminal goal.

To put progress on a shared scale, we report the native outcome (e.g., lines or points) and a normalized score for episode $e$ in game type $f$:
\begin{equation}
 q(e)=100\min\left(\frac{p(e)}{R_f},1\right),
\end{equation}
where $p(e)$ is peak progress, unaffected by later failure, and $R_f$ is a game type reference target: 3200 pixels (Mario), 85 food (Snake), 260 lines (Tetris), 130,000 points (Temple), 11,000 points (River), and 25 pipes (Bird). Targets are shared across models and conditions.

On this scale, a score of 100 means reaching the reference target, not necessarily winning the native game. Normalization provides a shared scale, not equal difficulty. A score for a game type is the mean over its tasks; Avg (or Average) is the unweighted mean across the six game types.

\subsection{Clock Sensitivity}
\label{sec:clock-controls}

We next examine whether an agent's performance with a paused game carries over to a continuously running game. We compare a fast decision model, Jev, with a slower reasoning LLM, GPT-6.1-sol, under matched tasks and different clocks.

\input{world-clock-results}
\input{figures/figure-clock-gap}

\section{TORSI: Two-Timescale Offline Recursive Self-Improvement}
\label{sec:torsi}

GPT-6.1-sol outperforms Jev in most paused games but falls behind under the live clock. Motivated by this observation, we propose TORSI, which combines their strengths by using GPT-6.1-sol to develop guidance offline and Jev to act on it during gameplay without teacher calls.

\subsection{Two-Timescale Agent}

\input{figures/figure-3}

TORSI follows four steps (Figure~\ref{fig:loops}): collect the student's failures on feedback tasks, practice corrections offline, propose guidance, and compare the frozen candidate with the current guidance on validation tasks. Accepted guidance starts the next round. Test tasks evaluate the resulting agent and never select updates.

For game type $f$ and round $k$, the student follows $\pi_F(\cdot;h_k^{(f)})$, where $h_k^{(f)}$ contains prompt instructions and conditional action rules; $h_0^{(f)}$ is the initial guidance. Only guidance changes; weights, observations, legal actions, physics, clocks, objectives, budgets, scoring, and splits remain fixed. The student chooses every action without teacher calls. The teacher explores saved feedback states offline; the deployed student cannot inspect future states.

\subsection{Learning from Feedback and Offline Practice}
\label{sec:offline-practice}

Each round starts by running the current student on feedback tasks under the live clock. The resulting trajectories $\mathcal D^{(f,k)}$ record observations, responses, response latency, action age, and outcomes, with screenshots aligned to requests.

Using these trajectories, the teacher examines timing errors, misunderstood mechanics, and failing plans. Trajectories show what happened, but not the outcomes of untried alternatives. \emph{Teacher practice} tests alternative action sequences from saved feedback states in paused or slowed environments. This access is restricted to offline learning and excludes test states.

\emph{Student practice} then checks whether the lesson works at the fast student's live latency. The student tries draft guidance on feedback tasks, and the teacher revises it using the resulting trajectories. Successful and failed trials both enter the practice evidence $\mathcal B^{(f,k)}$.

\subsection{Turning Practice into Reusable Guidance}

The teacher combines the current guidance with this feedback and practice evidence to propose revised guidance, using update procedure $T_k$ for round $k$:
\begin{equation}
 \widetilde h_{k+1}^{(f)}=T_k\big(h_k^{(f)},\mathcal D^{(f,k)},\mathcal B^{(f,k)}\big).
\end{equation}
A practiced action sequence fits a particular state; reusable guidance explains how to choose actions in other states. For example, a Mario instruction can specify releasing the jump button before initiating another jump, allowing the student to apply the rule whenever the relevant state occurs.

To deliver these rules to the student, guidance uses fixed fields for skill instructions, action-specific instructions, and conditional rules (Table~\ref{tab:updates}). Rules use current observation features, such as board height, visible hazards, and timing cues, and remain concise for action selection.

\subsection{Validation-Gated Rounds}
\label{sec:adoption}

To select an update, each candidate is frozen before evaluation on the game type's validation tasks. It replaces the current guidance only if its mean validation score is strictly higher; the current guidance stays unchanged for ties or rejected candidates. Adoption is decided separately for each game type using identical observation and action interfaces. Feedback supports learning, validation decides adoption, and test supports final reporting.

Subsequent rounds collect new feedback from the current student: adopted guidance changes its remaining failures, while rejection preserves the current guidance.

\section{Experiments}

We evaluate baseline models, TORSI improvement, alternative adaptation methods, and cross-model transfer. The clock controls appear in \S\ref{sec:clock-controls}. TORSI pairs Jev with GPT-6.1-sol (medium reasoning effort) as the offline teacher; Base denotes Jev with its initial guidance.

\setcounter{dbltopnumber}{3}
\begin{table*}[!t]
\centering\small
\setlength{\tabcolsep}{3pt}
\resizebox{\textwidth}{!}{%
\begin{tabular}{@{}llrrrrrrrrr@{}}
\toprule
Type & Model & Mario & Snake & Tetris & Temple & River & Bird & Average & Latency (s) & Cost (\$)\\
\midrule
\raisebox{-54pt}[0pt][0pt]{Decision} & GLiClass (0.4B) & 14.40 & 0.12 & 0.00 & 1.39 & 8.61 & 2.00 & 4.42 & 0.40 & ---\\
 & Clef-flash 9B & 17.66 & 0.24 & 0.19 & 4.05 & \textbf{11.43} & 12.40 & 7.66 & 0.31 & ---\\
 & Clef 27B & 18.57 & \textbf{22.71} & 5.38 & 5.09 & 9.51 & \textbf{36.80} & 16.34 & 0.65 & ---\\
 & StartLux 4B & 19.69 & 21.53 & 7.89 & 2.04 & 11.34 & 10.80 & 12.22 & 0.28 & ---\\
 & StartLux 27B & 14.52 & 2.35 & \textbf{33.12} & \textbf{37.16} & 3.37 & 9.20 & \textbf{16.62} & 0.82 & ---\\
 & JevK5 4B & \textbf{22.60} & 1.62 & 2.62 & 4.94 & 4.74 & 8.00 & 7.42 & 0.249 & ---\\
 & Open-Jev-2B & 5.08 & 0.24 & 0.15 & 1.47 & 2.96 & 0.00 & 1.65 & 0.413 & ---\\
 & mini-Jev & 20.00 & 0.12 & 0.00 & 0.63 & 3.31 & 0.00 & 4.01 & 0.467 & ---\\
 & decision-model-preview & 10.39 & 0.12 & 0.11 & 1.22 & 8.96 & 2.00 & 3.80 & 0.87 & ---\\
 & Jev & 21.78 & 14.98 & 17.73 & 10.86 & 8.63 & 7.07 & 13.51 & 0.52 & 0.0111\\
\midrule
\raisebox{-6pt}[0pt][0pt]{Language} & MiMo v2.6 Ultra Speed & 16.52 & 3.18 & 6.39 & 1.26 & 4.36 & 4.00 & 5.95 & 1.60 & 0.3173\\
 & DeepSeek v4.1 Flash & 9.47 & 0.12 & 1.42 & 0.70 & 0.55 & 2.40 & 2.44 & 7.72 & 0.0321\\
 & GPT-6.1-sol & 6.20 & 0.12 & 0.00 & 0.63 & 0.10 & 2.00 & 1.51 & 18.97 & 0.5109\\
\bottomrule
\end{tabular}}
\caption{Test scores without adaptation. Latency is the mean response latency $L_i$, averaged over game types. Cost is the mean API fee per task; dashes denote excluded local compute costs or unavailable API billing. GPT-6.1-sol uses medium reasoning effort; GLiClass~\citep{stepanov2025gliclass} uses the Instruct-Large v1.0 checkpoint. The decision-model-preview model is served by Alibaba Cloud.}
\label{tab:model-heldout}
\end{table*}

\begin{table*}[!t]
\centering\small
\setlength{\tabcolsep}{7pt}
\begin{tabular}{lrrrrrrr}
\toprule
 & \multicolumn{3}{c}{Validation} & \multicolumn{4}{c}{Test}\\
\cmidrule(lr){2-4}\cmidrule(lr){5-8}
Game type & Base & R1 & R2 & Base & R1 & R2 & Adopted\\
\midrule
Mario & 31.19 & \textbf{32.49} & 22.38 & 21.78 & \textbf{25.02} & 23.07 & \textbf{25.02}\\
Snake & 16.59 & 17.71 & \textbf{27.18} & 14.98 & 19.41 & \textbf{26.59} & \textbf{26.59}\\
Tetris & 16.54 & 23.73 & \textbf{24.96} & 17.73 & 18.76 & \textbf{27.77} & \textbf{27.77}\\
Temple & 3.65 & \textbf{47.04} & 26.29 & 10.86 & \textbf{24.90} & 23.10 & \textbf{24.90}\\
River & 20.35 & 24.60 & \textbf{27.74} & 8.63 & 13.71 & \textbf{24.87} & \textbf{24.87}\\
Bird & 10.00 & 18.00 & \textbf{19.20} & 7.07 & 14.27 & \textbf{16.00} & \textbf{16.00}\\
\midrule
Average & 16.38 & \textbf{27.26} & 24.63 & 13.51 & 19.35 & 23.57 & \textbf{24.19}\\
\bottomrule
\end{tabular}
\caption{Jev across two improvement rounds. Base uses initial guidance; R1 and R2 use round-1 and round-2 guidance. Each score is the mean of three runs. Adopted reports the final guidance selected by validation (\S\ref{sec:adoption}).}
\label{tab:results}
\label{tab:rsi-round2}
\end{table*}

\begin{table*}[!t]
\centering\small
\setlength{\tabcolsep}{3pt}
\resizebox{\textwidth}{!}{%
\begin{tabular}{@{}l*{7}{r}@{}}
\toprule
Method & Mario & Snake & Tetris & Temple & River & Bird & Avg\\
\midrule
Base & 31.19/21.78 & 16.59/14.98 & 16.54/17.73 & 3.65/10.86 & 20.35/8.63 & 10.00/7.07 & 16.38/13.51\\
Instr. only & 27.99/21.78 & 5.29/14.98 & 20.19/17.42 & 4.48/4.17 & 21.11/13.08 & 8.00/7.07 & 14.51/13.08\\
+ Teacher prac. & 28.48/21.78 & 8.47/14.98 & 17.54/19.96 & 9.48/10.07 & 20.51/10.47 & 10.80/7.60 & 15.88/14.14\\
+ Student prac. & 31.40/21.27 & 8.12/14.98 & 24.62/14.15 & 5.27/4.56 & 10.75/8.63 & 14.40/14.80 & 15.76/13.07\\
Full TORSI R1 & 32.49/25.02 & 17.71/19.41 & 23.73/18.76 & 47.04/24.90 & 24.60/13.71 & 18.00/14.27 & 27.26/19.35\\
\bottomrule
\end{tabular}%
}
\caption{One-round ablation of offline practice. Each cell reports validation (candidate) / test (adopted) scores. All variants generate teacher guidance from student failure trajectories. We independently ablate teacher and student practice. Adoption is determined per game type by validation improvement over Base; Base is used for test evaluation when a candidate is rejected.}
\label{tab:no-practice}
\end{table*}

\begin{table*}[t]
\centering\small
\setlength{\tabcolsep}{2.3pt}
\resizebox{\textwidth}{!}{%
\begin{tabular}{@{}lrrrrrrrrrrrr@{}}
\toprule
 & \multicolumn{7}{c}{Test score} & Jev & \multicolumn{3}{c}{GPT-6.1-sol medium} & Total\\
\cmidrule(lr){2-8}\cmidrule(lr){9-9}\cmidrule(lr){10-12}\cmidrule(lr){13-13}
Method & Mario & Snake & Tetris & Temple & River & Bird & Avg & Cost (\$) & Input (M) & Output (M) & Cost (\$) & Cost (\$)\\
\midrule
Base & 21.78\textsuperscript{\phantom{\(\uparrow\)}} & 14.98\textsuperscript{\phantom{\(\uparrow\)}} & 17.73\textsuperscript{\phantom{\(\uparrow\)}} & 10.86\textsuperscript{\phantom{\(\uparrow\)}} & 8.63\textsuperscript{\phantom{\(\uparrow\)}} & 7.07\textsuperscript{\phantom{\(\uparrow\)}} & 13.51\textsuperscript{\phantom{\(\uparrow\)}} & 0 & 0 & 0 & 0 & 0.00\\
ACE & --- & --- & --- & 16.25\textsuperscript{\(\uparrow\)} & 2.16\textsuperscript{\(\downarrow\)} & 6.40\textsuperscript{\(\downarrow\)} & 13.22\textsuperscript{\(\downarrow\)} & 4.30 & 8.5417 & 0.2672 & 19.76 & 24.05\\
AHE & --- & --- & 18.81\textsuperscript{\(\uparrow\)} & 18.38\textsuperscript{\(\uparrow\)} & 18.55\textsuperscript{\(\uparrow\)} & --- & 16.59\textsuperscript{\(\uparrow\)} & 3.00 & 10.5587 & 0.3375 & 24.49 & 27.49\\
GEPA & --- & 17.30\textsuperscript{\(\uparrow\)} & 26.65\textsuperscript{\(\uparrow\)} & 5.69\textsuperscript{\(\downarrow\)} & --- & --- & 14.52\textsuperscript{\(\uparrow\)} & 6.46 & 6.6958 & 0.1144 & 14.54 & 20.99\\
\midrule
\textbf{TORSI (ours)} & \textbf{25.02}\textsuperscript{\(\uparrow\)} & \textbf{26.59}\textsuperscript{\(\uparrow\)} & \textbf{27.77}\textsuperscript{\(\uparrow\)} & \textbf{24.90}\textsuperscript{\(\uparrow\)} & \textbf{24.87}\textsuperscript{\(\uparrow\)} & \textbf{16.00}\textsuperscript{\(\uparrow\)} & \textbf{24.19}\textsuperscript{\(\uparrow\)} & 1.46 & 11.2187 & 0.0490 & 22.93 & 24.39\\
\bottomrule
\end{tabular}}
\caption{Final adopted test scores and total API cost of adaptation. Dashes indicate no adopted update; Avg includes Base scores for those game types. Input and output counts are millions of teacher tokens; output includes reasoning tokens. Appendix~\ref{app:cost} details the cost accounting.}
\label{tab:adaptation-comparison}
\end{table*}

\begin{table}[!t]
\centering\small
\setlength{\tabcolsep}{5pt}
\begin{tabular}{@{}llrrrrrrr@{}}
\toprule
Target model & & Mario & Snake & Tetris & Temple & River & Bird & Average\\
\midrule
StartLux 4B & Before & 19.69 & 21.53 & 7.89 & 2.04 & 11.34 & 10.80 & 12.22\\
 & After & \textbf{22.17} & \textbf{21.77} & \textbf{13.04} & \textbf{64.51} & \textbf{16.00} & \textbf{12.80} & \textbf{25.05}\\
\addlinespace
StartLux 27B & Before & 14.52 & 2.35 & \textbf{33.12} & \textbf{37.16} & 3.37 & 9.20 & 16.62\\
 & After & \textbf{16.87} & \textbf{19.53} & 24.08 & 35.27 & \textbf{14.40} & \textbf{12.40} & \textbf{20.43}\\
\addlinespace
MiMo v2.6 Ultra Speed & Before & 16.52 & 3.18 & 6.39 & 1.26 & 4.36 & 4.00 & 5.95\\
 & After & \textbf{17.86} & \textbf{6.35} & \textbf{7.46} & \textbf{1.48} & \textbf{5.45} & \textbf{5.20} & \textbf{7.30}\\
\bottomrule
\end{tabular}
\caption{Transfer of frozen Jev-derived guidance to StartLux and MiMo models (test scores). Bold marks the higher of each Before/After pair.}
\label{tab:model-transfer}
\end{table}

\subsection{Benchmark Evaluation}
\label{sec:model-heldout}

For the baseline comparison, models receive the same state information and legal choices under identical clocks and budgets. Decision models score choices; LLMs return a choice. Table~\ref{tab:model-heldout} reports scores on ten test tasks per game type, mean response latency, and API cost per task.

Decision-model mean scores range from 1.65 to 16.62, with mean response latencies below one second. GPT-6.1-sol scores 1.51 with a latency of 18.97 seconds, while MiMo v2.6 Ultra Speed and DeepSeek v4.1 Flash, both promoted for faster inference \citep{xiaomimimo2026release,deepseek2026flash}, score 5.95 and 2.44 with latencies of 1.60 and 7.72 seconds, respectively. Even these speed-oriented LLMs achieve limited progress under these real-time evaluation conditions.

\subsection{TORSI Improvement Rounds}
\label{sec:jev-adaptation}

After two rounds, the validation-selected guidance raises Jev's mean test score from 13.51 to 24.19, a gain of 10.68 points, with improvements in all six game types (Table~\ref{tab:results}).

\paragraph{Round selection.} R1 and R2 denote guidance from the first and second rounds. R1 improves test scores in all six game types. R2 improves both splits for Snake, Tetris, River, and Bird, but fails the validation gate for Mario and Temple. Adopted therefore uses R1 for those two game types and R2 for the other four, selected by validation rather than test performance. Scores average three runs.

An illustrated Mario example in Appendix~\ref{app:evidence} (Figure~\ref{fig:reasoning-guidance}) follows the first round from feedback and offline practice to held-out gameplay.

\paragraph{Roles of offline practice.} To understand how practice contributes to these gains, we compare variants with and without teacher and student practice. All one-round variants in Table~\ref{tab:no-practice} generate guidance from failures. Without practice, the adopted mean test score is 13.08, versus Base's 13.51. Teacher-only and student-only practice give 14.14 and 13.07; combining both gives 19.35.

Testing alternatives and checking their live execution are complementary here, although practice budgets are not matched across variants.

\subsection{Comparison with Other Adaptation Methods}
\label{sec:adaptation-comparison}

For comparison with TORSI, we adapt ACE, AHE, and GEPA to the same Jev student, offline teacher, task splits, live-clock evaluation, and validation adoption rule \citep{zhang2025ace,lin2026ahe,agrawal2025gepa}. Editable components and search configurations differ (Appendix~\ref{app:adaptation-protocol}), so this comparison evaluates the adapted configurations without matched budgets.

In this comparison, TORSI achieves the highest adopted mean test score, 24.19, compared with 13.22 for ACE, 16.59 for AHE, and 14.52 for GEPA (Table~\ref{tab:adaptation-comparison}). Where no update is adopted, the table uses Base scores in the average. Total API expenditure over the update process is \$24.39 for TORSI and \$20.99--\$27.49 for the alternatives. These total adaptation costs differ from per-task API fees and response latency.

\subsection{Cross-Model Generalization}
\label{sec:model-transfer}

We evaluate whether guidance developed with Jev generalizes to other models without further adaptation. We apply the final guidance unchanged, keeping the same tasks and clocks (Table~\ref{tab:model-transfer}). MiMo uses an action-selection adapter that returns a valid choice from the provided list. StartLux 4B's mean score rises from 12.22 to 25.05, StartLux 27B's from 16.62 to 20.43, and MiMo's mean score from 5.95 to 7.30.

Transfer gains vary across models: StartLux 4B and MiMo improve in all six game types, while StartLux 27B improves in four but regresses in Tetris and Temple. These tests use the same game types and do not evaluate guidance transfer to previously unseen environments.

\section{Conclusion}

Real-time intelligence requires both capable decisions and timely action. RealtimeBench provides a shared evaluation framework for studying this requirement and comparing future approaches. TORSI demonstrates one such approach: a powerful LLM reasons and practices offline, turning its experience into reusable guidance that the fast Jev actor applies during live interaction. Improvements across all six game types, together with gains from transferring the guidance to other models, suggest that this combination can extend beyond a single actor. This points toward a promising paradigm for real-time intelligence, in which stronger LLMs continually improve fast actors without requiring those actors to wait for teacher inference during live gameplay.

\section*{Limitations}

\paragraph{Structured observations.} RealtimeBench does not evaluate actors operating directly from raw visual observations. Actors receive structured state descriptions, so the benchmark omits errors and delays introduced by visual perception. This limits the applicability of our findings to end-to-end visual agents. Nevertheless, all evaluated decision models and LLMs receive the same state information and legal action choices, so the reported comparisons remain informative about decision-making under inference delay within this shared interface.

\paragraph{Access to offline practice.} TORSI relies on saved states or resettable environments for teacher practice. Such access may be unavailable in physical or irreversible interactions, limiting direct application beyond simulation. Our experiments therefore establish the method’s effectiveness where offline practice is available; extending it to other settings would require an appropriate mechanism for testing proposed corrections.

\paragraph{Generalization and experimental scope.} Our experiments cover six game types and two improvement rounds, with guidance developed separately for each game type. Cross-model transfer retains the same environments and produces some game-specific regressions; it does not establish transfer to unseen environments or sustained improvement over many rounds. The observed gains nevertheless provide evidence that guidance can be reused across the evaluated actors without retuning.

\section*{Ethical Considerations}

More capable and responsive agents could make automated interaction easier to scale, creating opportunities for misuse such as unfair gameplay or unauthorized actions on digital services. Teacher-generated guidance may carry over errors or harmful instructions from the underlying models, while a fast actor can apply them repeatedly before human intervention. Reusing the same guidance across models could further spread these risks.

\bibliographystyle{colm2026_conference}
\bibliography{references}

\appendix
\raggedbottom
\urlstyle{same}
\setcounter{dbltopnumber}{3}
\renewcommand{\dbltopfraction}{0.9}
\renewcommand{\textfraction}{0.08}
\renewcommand{\dblfloatpagefraction}{0.7}
\input{appendix-layout}

\section{Benchmark and Guidance Details}
\label{app:evidence}
\input{tables/benchmark-comparison}
\AppendixTaskSettings
\AppendixActionInterfaces
\AppendixGuidanceLessons
\input{figures/figure-mario-practice}
\input{figures/teacher-reasoning-examples}

\paragraph{Intended use.} RealtimeBench is intended for research and evaluation of latency-sensitive agents in controlled game environments. TORSI is intended to study how offline teacher guidance improves real-time control within these environments.

\paragraph{Underlying implementations.} Mario uses gym-super-mario-bros (9.1.0) and nes-py (9.0.1), whose emulator core derives from SimpleNES. Temple uses Three.js (0.180.0), while River and Bird use Pygame-based implementations. Snake and Tetris use deterministic Python and JavaScript game engines, respectively. Software sources are listed in Appendix~\ref{app:software-sources}.

\paragraph{Benchmark-specific task construction.} We construct RealtimeBench through explicit task specifications over six game families. Each task fixes the game version, seed or stage, difficulty, observation and action interfaces, and episode budget. Task configurations span platform layouts, board geometry and hazards, falling-piece sequences and gravity, running speed, starting lives, and pipe-gap width. These specifications support paired evaluations under different clocks while keeping task identity and game physics fixed. Feedback, validation, and test tasks serve distinct roles in experience collection, update selection, and final evaluation.

\paragraph{Live-clock adaptation and timing-aware interfaces.} Our execution framework couples asynchronous model inference with continuously advancing game state and game-specific action semantics. It records observation, response, and action-application times, allowing response latency and action age to be measured separately. The adapters connect structured observations to executable controls, including timestamped button plans, scheduled turns, piece placements, steering decisions, and flap schedules. For Snake, our adapter extends the underlying engine with configurable task presets, current-grid topology features, response-delay information, and joint direction-and-turn-timing decisions. Turn timing is anchored to the observation tick, making the consequences of inference delay explicit in action execution.

\paragraph{Evaluation and traceability.} We define a shared progress-based evaluation procedure while retaining each game's native outcome. Scores use game-specific reference targets that remain fixed across models and clock conditions, followed by task-level and game-level aggregation. Recorded states and action logs support checking episode outcomes. The task manifest identifies underlying game sources and revisions, while Tables~\ref{tab:suite}--\ref{tab:observation-action-interfaces} document task settings, clocks, observations, and action interfaces. The release includes task definitions, per-episode outcomes, frozen agent specifications, reusable guidance, and replay logs, making the evaluated system and its execution conditions explicit.

Table~\ref{tab:updates} and Figure~\ref{fig:guidance-types} illustrate guidance formats and representative lessons.

Figure~\ref{fig:reasoning-guidance} uses native Mario scenes and poses replayed from archived feedback, teacher practice, and held-out episodes. Poses are aligned using recorded camera positions and enlarged for visibility.

Two improving student-practice examples are shown; the archived three-task practice screen left the guidance unchanged. The held-out Before/After panels show the initial student (Base) and the student with round-1 guidance (R1) on the same task. These examples illustrate the full guidance update rather than isolate the effect of an individual rule.

\section{Additional Experimental Results}
\label{app:additional-results}
\AppendixAdaptationMethods
\AppendixAdaptationSearch
\input{tables/clock-controls}
\AppendixResponseLatency
\AppendixApiCost

\subsection{Adaptation Comparison Protocol}
\label{app:adaptation-protocol}

All methods use Jev as the real-time student and GPT-6.1-sol with medium reasoning effort as the offline teacher. Each game type has ten feedback, ten validation, and ten test tasks. Game physics, clocks, scoring, observations, and legal action interfaces remain fixed. The teacher is never queried during scored play. A frozen candidate is adopted only if its mean validation score strictly exceeds that of the current agent; test tasks are used solely for final evaluation.

Table~\ref{tab:adaptation-protocol} summarizes how each method's optimization mechanism is adapted to the RealtimeBench setting, and Table~\ref{tab:adaptation-search} lists the corresponding search configurations and stopping conditions. All methods use repeated task-level rollouts on the same ten feedback tasks as their interaction unit.

Episode budgets count student executions of feedback tasks, including repeated executions of the same task; validation and test evaluations are accounted for separately.

\subsection{Response Latency}
\label{app:latency}

\paragraph{Model execution.} Our local inference hardware consists of NVIDIA L20Y GPUs. All local models run on a single GPU without tensor or pipeline parallelism, e.g., GLiClass (0.4B), Open-Jev-2B (2B), JevK5 (4B), Clef-flash (9B), Clef (27B), and StartLux (4B and 27B). Local inference incurs no API fees; its compute cost is the wall-clock response time reported in Table~\ref{tab:latency}, and no additional compute budget was imposed on local runs. Among the API-based models, DeepSeek v4.1 Flash is queried with thinking mode disabled to reduce response latency, and MiMo v2.6 Ultra Speed exposes no thinking interface, so it is called through the official API with the provider's default decoding settings.

Decision models have median response times below two seconds in every game type. In Snake, all of them except StartLux 27B (0.86~s) stay below the 667~ms tick. MiMo has a low median but a long tail, reaching 13.64~s at P95 in Mario. GPT-6.1-sol's game type medians range from 5 to 31~s, during which the live environment continues evolving.

\subsection{Token Usage and Computational Cost}
\label{app:cost}

Table~\ref{tab:api-cost} reports token usage and API cost for the recorded task runs. In Table~\ref{tab:adaptation-comparison}, token counts are in millions (with reasoning tokens included in output) and API costs are totals over the full update process. Jev costs are estimated at \$0.042/M input tokens, with free output; teacher costs use \$2/M input tokens and \$10/M output tokens. GPUs were used only to serve open-source models during evaluation, totaling 46.47 GPU-hours.

\subsection{Model Sources} Official model cards document the evaluated GLiClass checkpoint,\footnote{\href{https://huggingface.co/knowledgator/gliclass-instruct-large-v1.0}{\nolinkurl{Hugging Face: knowledgator/gliclass-instruct-large-v1.0}}} JevK5,\footnote{\href{https://huggingface.co/alibiserikbay/JevK5}{\nolinkurl{Hugging Face: alibiserikbay/JevK5}}} Open-Jev-2B,\footnote{\href{https://huggingface.co/ZefanCai/Open-Jev-2B}{\nolinkurl{Hugging Face: ZefanCai/Open-Jev-2B}}} mini-Jev,\footnote{\href{https://huggingface.co/samatv256/mini-Jev}{\nolinkurl{Hugging Face: samatv256/mini-Jev}}} and StartLux 27B.\footnote{\href{https://huggingface.co/startlux-models/StartLux-Decision-27B}{\nolinkurl{Hugging Face: startlux-models/StartLux-Decision-27B}}} Official API documentation lists MiMo v2.6 Ultra Speed\footnote{\href{https://mimo.mi.com/docs/en-US/api/chat/openai-api}{\nolinkurl{mimo.mi.com/docs/en-US/api/chat/openai-api}}} and DeepSeek v4.1 Flash.\footnote{\href{https://api-docs.deepseek.com/quick_start/pricing/}{\nolinkurl{api-docs.deepseek.com/quick_start/pricing/}}}

\section{Software Sources}
\label{app:software-sources}

\textbf{Mario emulator stack.} Mario uses Christian Kauten's gym-super-mario-bros (9.1.0)\footnote{\href{https://github.com/Kautenja/gym-super-mario-bros}{\nolinkurl{GitHub: Kautenja/gym-super-mario-bros}}} and nes-py (9.0.1),\footnote{\href{https://github.com/Kautenja/nes-py}{\nolinkurl{GitHub: Kautenja/nes-py}}} with an emulator core derived from Amish Kumar Naidu and contributors' SimpleNES.\footnote{\href{https://github.com/amhndu/SimpleNES}{\nolinkurl{GitHub: amhndu/SimpleNES}}}

\textbf{Rendering frameworks.} Temple uses Three.js (0.180.0, r180).\footnote{\href{https://github.com/mrdoob/three.js}{\nolinkurl{GitHub: mrdoob/three.js}}} River and Bird use Pygame-based implementations, with sources maintained by the Pygame\footnote{\href{https://github.com/pygame/pygame}{\nolinkurl{GitHub: pygame/pygame}}} and Pygame Community Edition\footnote{\href{https://github.com/pygame-community/pygame-ce}{\nolinkurl{GitHub: pygame-community/pygame-ce}}} contributor communities.

\textbf{Game implementations.} The original Super Mario Bros. game is by Nintendo.\footnote{\href{https://www.nintendo.com/en-gb/Games/NES/Super-Mario-Bros-803853.html}{Nintendo: Super Mario Bros.}} Snake adapts iammusham's jev-snake (revision \texttt{a935107}),\footnote{\href{https://github.com/iammusham/jev-snake}{\nolinkurl{GitHub: iammusham/jev-snake}}} and Tetris adapts trungdq88's jev-tetris (\texttt{a0b411c}).\footnote{\href{https://github.com/trungdq88/jev-tetris}{\nolinkurl{GitHub: trungdq88/jev-tetris}}} Temple adapts Mukhsin Mukhtorov's Night Relay (\texttt{1e53984}).\footnote{\href{https://github.com/Mukhsin0508/night-relay}{\nolinkurl{GitHub: Mukhsin0508/night-relay}}} River uses Ashkan Aazami's river-run-typesafe (\texttt{8cf0429}),\footnote{\href{https://github.com/ashaazami/river-run-typesafe}{\nolinkurl{GitHub: ashaazami/river-run-typesafe}}} and Bird uses leftspace89's JevBird (\texttt{1e5fd13}).\footnote{\href{https://github.com/leftspace89/JevBird}{\nolinkurl{GitHub: leftspace89/JevBird}}}

\end{document}

%% file: figures/figure-1.tex
\begin{figure*}[t!]
\centering
\includegraphics[width=\textwidth]{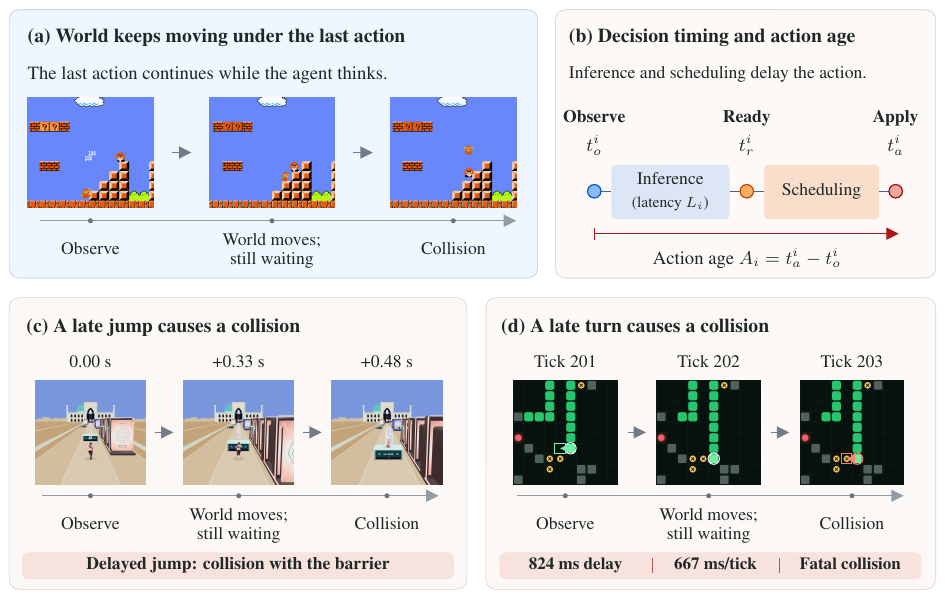}
\caption{Actions can become unsafe during inference. (a) Mario continues under the last action and collides with an enemy while the response is pending. (b) Action age is the time from observation to application, including inference and scheduling delays. (c) A delayed jump in Temple leads to a barrier collision; frames show observation, continued motion, and collision at 0.00, +0.33, and +0.48\,s. (d) LEFT is safe at Snake's observed tick 201, but the 824\,ms response exceeds a 667\,ms tick. Snake moves down before LEFT takes effect, so the turn hits a fatal decoy at tick 203.}
\label{fig:realtime-example}
\end{figure*}

%% file: figures/figure-2.tex
\begin{figure*}[!t]
\centering
\includegraphics[width=\textwidth]{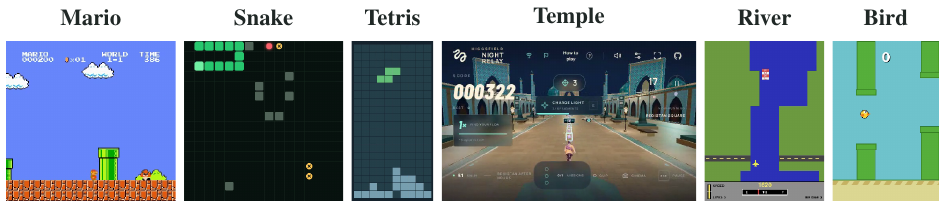}
\caption{The six RealtimeBench game types, shown with frames from recorded episodes (Tetris is rendered from recorded board states).}
\label{fig:overview}
\end{figure*}

%% file: figures/figure-base-harness.tex
\begin{figure*}[!t]
\centering
\includegraphics[width=\textwidth]{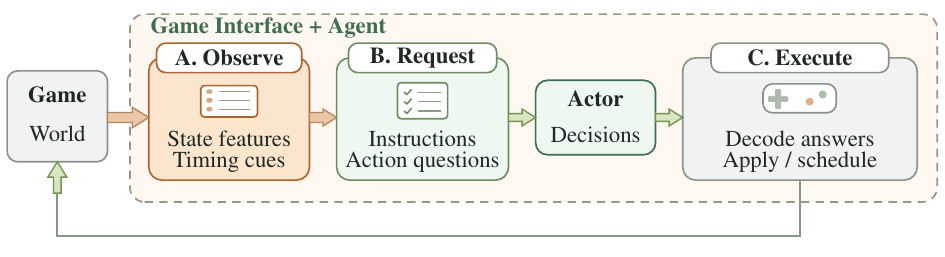}
\caption{The live action loop. The harness turns current game state into structured features and action questions, queries the agent, and applies decoded actions while the world keeps moving. Action requests include instructions and legal choices based on the current observation.}
\label{fig:base-harness}
\end{figure*}

%% file: world-clock-results.tex
Both models use unchanged instructions on the ten test tasks per game type; GPT-6.1-sol uses medium reasoning effort. We evaluate four clocks: Paused, which freezes the game during inference; Current, the declared live clock; and 1.5$\times$ and 2$\times$ faster clocks. Seeds, instructions, legal actions, objectives, and budgets are fixed; pacing and the clock fields shown to the agent change together.

\paragraph{Pausing changes the comparison.} GPT-6.1-sol outperforms Jev in five game types when paused, but under the live clock its mean score is 1.51 versus Jev's 13.51 (Figure~\ref{fig:clock-gap}). GPT-6.1-sol scores lower live than paused in all six game types; Jev does so in five, with Bird as the exception. Within-model comparisons reveal the outcome cost of allowing the game to evolve during inference.

\paragraph{Sensitivity varies by game type.} Faster clocks generally reduce scores, but individual changes are not monotonic. Bird's relative robustness at moderate speeds may reflect its interface, which executes flap schedules planned in advance; this is an interpretation rather than an isolated causal test. Temple drops sharply from Paused to Current. Together, the controls show why an agent that reasons well with a paused clock may still need better decisions under live timing.

%% file: figures/figure-clock-gap.tex
\begin{figure*}[t!]
\centering
\includegraphics[width=\textwidth]{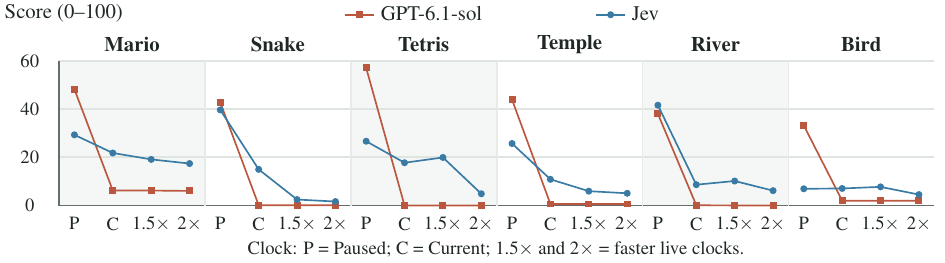}
\caption{Clock sensitivity without adaptation. Each group compares GPT-6.1-sol (medium reasoning effort) and Jev under Paused, Current, 1.5$\times$, and 2$\times$ clocks. Scores are mean normalized progress over ten test tasks (0--100).}
\label{fig:clock-gap}
\end{figure*}

%% file: figures/figure-3.tex
\begin{figure*}[!t]
\centering
\includegraphics[width=\textwidth]{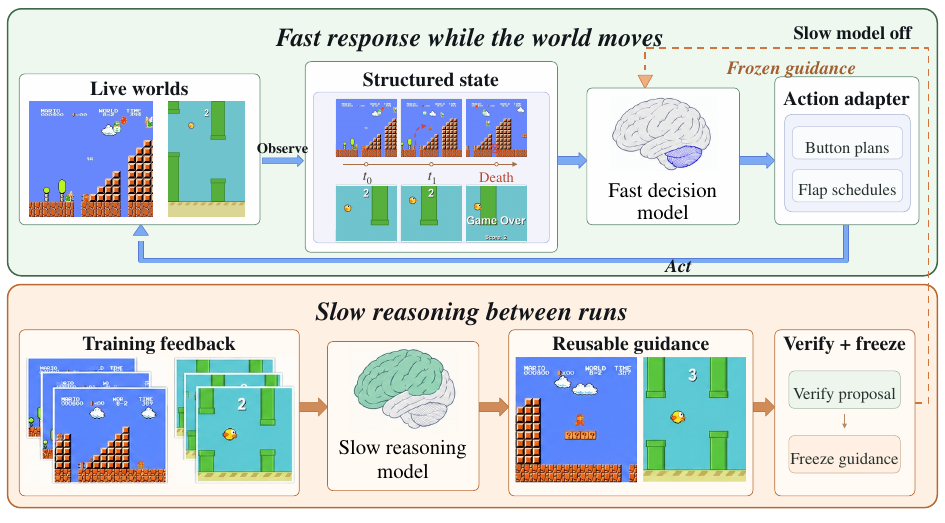}
\caption{TORSI separates live acting from offline learning. The student acts alone during play; the teacher turns feedback and offline practice into guidance between episodes. Frozen candidates are selected on validation tasks and evaluated on held-out test tasks.}
\label{fig:loops}
\end{figure*}

%% file: appendix-layout.tex
\newcommand{\AppendixTaskSettings}{%
\begin{table*}[!tb]
\centering\small
\setlength{\tabcolsep}{4pt}
\resizebox{\textwidth}{!}{%
\begin{tabular}{llll}
\toprule
Game type & Task variation & Live clock & Progress measure\\
\midrule
Mario & ROM stages grouped by world and mechanics & 10 FPS, three lives & Furthest $x$ position\\
Snake & Board size, obstacles, decoy food & 1.5 moves/s & Food collected\\
Tetris & Seeded pieces, gravity levels & 450--200 ms per row & Lines cleared\\
Temple & Seeded route, game-speed levels & 0.30--0.70$\times$ original speed & Points\\
River & Seeded river layout, 1--5 starting lives & 60 FPS & Points\\
Bird & Seeded pipes, 200--120 px gaps & 20 FPS & Pipes cleared\\
\bottomrule
\end{tabular}}
\caption{Task settings for each game type. Each task additionally fixes its game version, action interface, and episode budget.}\label{tab:suite}
\end{table*}
}

\newcommand{\AppendixActionInterfaces}{%
\begin{table*}[!tb]
\centering\small
\setlength{\tabcolsep}{4pt}
\renewcommand{\arraystretch}{1.12}
\begin{tabular}{@{}>{\raggedright\arraybackslash}p{0.09\textwidth}>{\raggedright\arraybackslash}p{0.34\textwidth}>{\raggedright\arraybackslash}p{0.52\textwidth}@{}}
\toprule
Game type & Observation & Action interface and execution\\
\midrule
Mario & Position, terrain, enemies, jump state, and timing features extracted from RAM. & Button plans over future time slots, with explicit short-jump, long-jump, and A-release semantics.\\
\addlinespace[3pt]
Snake & Board, body, food, obstacles, and timing features. & Direction and delay in ticks; repeated choices of the same direction preserve the earlier pending turn.\\
\addlinespace[3pt]
Tetris & Board, current and next pieces, candidate placements, and geometric features. & Joint rotation/column choice; reachability checks when the response arrives and an optional re-query for an unreachable placement.\\
\addlinespace[3pt]
Temple & Lane, pose, visible hazards, and estimated time to contact. & Wait, move left or right, jump, slide, or activate Pulse.\\
\addlinespace[3pt]
River & Visible river channels, enemies, fuel, and plane state. & Lane, fire, and throttle decisions; a steering converter maps model outputs to continuous steering.\\
\addlinespace[3pt]
Bird & Height, velocity, visible pipes, and a projection of the currently held flap schedule. & Eight complete flap schedules; the selected schedule continues between responses.\\
\bottomrule
\end{tabular}
\caption{Observations and action interfaces for the six game types.}
\label{tab:observation-action-interfaces}
\end{table*}
}

\newcommand{\AppendixGuidanceLessons}{%
\begin{table*}[!tb]
\centering\small
\setlength{\tabcolsep}{4pt}
\begin{tabular}{@{}p{0.08\textwidth}p{0.20\textwidth}p{0.33\textwidth}p{0.33\textwidth}@{}}
\toprule
Game type & Guidance format & Round-1 lesson & Round-2 lesson\\
\midrule
Mario & Skill instructions; action and timing rules & Skill guidance learned from visual practice: release before a new jump, hold the jump button for height when ceiling clearance allows, land on support, account for action age, and keep moving after landing. & $^{\dagger}$After a supported landing, continue along the ground when no visible hazard requires a jump. Release the jump button after landing before starting the next jump. Time takeoff using nearby hazards and landing surfaces.\\
\addlinespace[3pt]
Tetris & Skill instructions; conditional rules; re-query on unreachable placement & On high boards or outdated observations, a conditional rule adds concise geometry guidance and allows one re-query when the chosen placement is unreachable. & Below height 17, prefer a line clear only when it also lowers peak height and adds at most one new buried cell relative to visible alternatives.\\
\addlinespace[3pt]
Snake & Direction and turn-timing instructions & Safety, food-seeking, and turn-timing guidance that balances immediate progress with a safe next move. & Check the next turn as well as the immediate move. Keep an escape route rather than chase food into a dead end near the board edge or a one-cell corridor.\\
\addlinespace[3pt]
Temple & Instructions for specific action choices & Revised jump, slide, lane-change, and special-action conditions based on pose, hazards, and expected action age. & $^{\dagger}$Move to a clear lane early. After each step, recheck actual lane position and visible hazards, then continue toward the clear destination.\\
\addlinespace[3pt]
River & Fire, lane, and throttle instructions & Revised fuel, shooting, open-water, and steering guidance. & Hold the lane while firing at an aligned stationary target. Maintain the route through the next bridge or enemy, and shoot moving targets when they enter the line of fire.\\
\addlinespace[3pt]
Bird & Flap-schedule instructions & Adjust flap timing to reach a safe height before entering a pipe gap and stay within the gap while passing through it. & Flap immediately if waiting would leave the bird below the pipe gap at entry; if it is too high, delay flapping to descend. Start correcting height early and maintain a safe height until the bird clears the pipe.\\
\bottomrule
\end{tabular}
\caption{Formats used to deliver teacher guidance and representative lessons from both improvement rounds. $^{\dagger}$ marks R1 or R2 candidate lessons rejected by the validation gate.}\label{tab:updates}
\end{table*}
}

\newcommand{\AppendixAdaptationMethods}{%
\begin{table*}[!tb]
\centering
\small
\renewcommand{\arraystretch}{1.12}
\setlength{\tabcolsep}{4pt}
\resizebox{\textwidth}{!}{%
\begin{tabular}{@{}>{\raggedright\arraybackslash}p{\dimexpr0.100\textwidth-2\tabcolsep\relax}>{\raggedright\arraybackslash}p{\dimexpr0.196\textwidth-2\tabcolsep\relax}>{\raggedright\arraybackslash}p{\dimexpr0.272\textwidth-2\tabcolsep\relax}>{\raggedright\arraybackslash}p{\dimexpr0.157\textwidth-2\tabcolsep\relax}>{\raggedright\arraybackslash}p{\dimexpr0.275\textwidth-2\tabcolsep\relax}@{}}
\toprule
Method & Core optimization mechanism & RealtimeBench adaptation & Editable components & Internal optimization and selection\\
\midrule
\textbf{TORSI} & Offline practice, guidance distillation, and recursive improvement. & Both rounds include feedback analysis, teacher practice, student guidance checks from saved states, and guidance updates. & Guidance and conditional action rules for each game type. & Successful and failed practice branches inform a guidance proposal, which is frozen for external validation.\\
\addlinespace
\textbf{ACE} & Generator--Reflector--Curator collaboration with incremental playbook refinement. & Game trajectories support reflection, student regeneration, and localized curation. Stable entry identifiers and attribution counters are preserved; the curator merges redundant entries. & Structured context playbook. & Incremental updates, reflection-driven regeneration, and multi-epoch refinement produce the final playbook.\\
\addlinespace
\textbf{AHE} & Execution-evidence-driven component evolution, effect prediction, and verification with rollback. & Layered trajectory diagnosis supports component edits and outcome checks. Editable components are restricted to the game interface. & Guidance, question instructions, memory, and conditional middleware. & Component edits are evaluated on feedback tasks and kept when mean feedback scores improve; otherwise, they are rolled back.\\
\addlinespace
\textbf{GEPA} & Reflective text optimization with Pareto selection and candidate merging. & The official optimizer uses a real-time adapter to evolve guidance and action-question text from game trajectories. & Guidance and action-question text. & The official search maintains a Pareto candidate pool, performs reflective mutations and merges, and returns its selected candidate.\\
\bottomrule
\end{tabular}}
\caption{Method-specific adaptations to RealtimeBench.}
\label{tab:adaptation-protocol}
\end{table*}
}

\newcommand{\AppendixAdaptationSearch}{%
\begin{table*}[!tb]
\centering
\small
\renewcommand{\arraystretch}{1.12}
\setlength{\tabcolsep}{4pt}
\begin{tabular}{@{}>{\raggedright\arraybackslash}p{\dimexpr0.100\textwidth-2\tabcolsep\relax}>{\raggedright\arraybackslash}p{\dimexpr0.900\textwidth-2\tabcolsep\relax}@{}}
\toprule
Method & Search configuration and stopping conditions\\
\midrule
\textbf{TORSI} & Two improvement rounds, each using ten feedback tasks together with teacher and student practice.\\
\addlinespace
\textbf{ACE} & Up to 60 feedback episodes per game type per search pass, five epochs, and two reflection-driven regenerations per task, subject to the remaining episode allowance.\\
\addlinespace
\textbf{AHE} & Up to 60 feedback episodes per game type per search pass and ten evolution iterations; stops after two consecutive iterations without an accepted edit.\\
\addlinespace
\textbf{GEPA} & Up to 60 feedback episodes per game type per search pass, controlled by the episode allowance and stopping callbacks. The adapted configuration uses five feedback tasks for reflection and five for Pareto evaluation.\\
\bottomrule
\end{tabular}
\caption{Search configurations for each adaptation method.}
\label{tab:adaptation-search}
\end{table*}
}

\newcommand{\AppendixResponseLatency}{%
\begin{table*}[!tb]
\centering\small
\setlength{\tabcolsep}{3.5pt}
\resizebox{\textwidth}{!}{%
\begin{tabular}{@{}lrrrrrrrrrrrr@{}}
\toprule
 & \multicolumn{2}{c}{Mario} & \multicolumn{2}{c}{Snake} & \multicolumn{2}{c}{Tetris} & \multicolumn{2}{c}{Temple} & \multicolumn{2}{c}{River} & \multicolumn{2}{c}{Bird}\\
\cmidrule(lr){2-3}\cmidrule(lr){4-5}\cmidrule(lr){6-7}\cmidrule(lr){8-9}\cmidrule(lr){10-11}\cmidrule(lr){12-13}
Model & P50 & P95 & P50 & P95 & P50 & P95 & P50 & P95 & P50 & P95 & P50 & P95\\
\midrule
Clef-flash 9B & 0.56 & 0.69 & 0.30 & 0.37 & 0.24 & 0.30 & 0.23 & 0.28 & 0.26 & 0.33 & 0.22 & 0.28\\
Clef 27B & 1.45 & 2.35 & 0.58 & 0.63 & 0.44 & 0.61 & 0.43 & 0.49 & 0.43 & 0.52 & 0.41 & 0.46\\
GLiClass (0.4B) & 1.80 & 3.10 & 0.22 & 0.25 & 0.08 & 0.17 & 0.07 & 0.11 & 0.10 & 0.20 & 0.09 & 0.09\\
Jev & 1.02 & 2.63 & 0.33 & 0.44 & 0.26 & 0.47 & 0.43 & 0.87 & 0.35 & 0.50 & 0.37 & 0.61\\
JevK5 4B & 0.690 & 1.291 & 0.187 & 0.195 & 0.211 & 0.325 & 0.085 & 0.126 & 0.147 & 0.206 & 0.087 & 0.091\\
Open-Jev-2B & 0.763 & 1.280 & 0.447 & 0.574 & 0.223 & 0.683 & 0.186 & 0.227 & 0.529 & 0.645 & 0.182 & 0.231\\
mini-Jev & 1.105 & 1.975 & 0.351 & 0.445 & 0.286 & 0.412 & 0.179 & 0.248 & 0.517 & 0.627 & 0.190 & 0.252\\
StartLux 4B & 0.61 & 0.91 & 0.24 & 0.42 & 0.11 & 0.28 & 0.08 & 0.34 & 0.22 & 0.49 & 0.28 & 0.31\\
StartLux 27B & 1.65 & 2.73 & 0.86 & 0.98 & 0.47 & 1.08 & 0.34 & 0.35 & 0.86 & 1.21 & 0.48 & 0.51\\
decision-model-preview & 0.647 & 1.620 & 0.530 & 6.537 & 0.318 & 3.746 & 0.260 & 1.483 & 0.171 & 0.649 & 0.264 & 1.281\\
MiMo v2.6 Ultra Speed & 0.88 & 13.64 & 0.55 & 1.39 & 0.55 & 1.39 & 0.60 & 7.93 & 0.54 & 1.19 & 0.56 & 1.48\\
DeepSeek v4.1 Flash & 2.90 & 8.54 & 3.27 & 36.79 & 2.61 & 3.79 & 5.43 & 49.67 & 2.98 & 32.75 & 2.66 & 4.33\\
GPT-6.1-sol & 25.11 & 78.90 & 14.03 & 34.70 & 31.06 & 86.08 & 10.08 & 14.27 & 9.30 & 14.29 & 5.49 & 13.52\\
\bottomrule
\end{tabular}}
\caption{Response latency $L_i$ in seconds: median (P50) and 95th percentile (P95) of wall-clock time from observation to response, including network time and all decision questions in a request.}
\label{tab:latency}
\end{table*}
}

\newcommand{\AppendixApiCost}{%
\begin{table*}[!tb]
\centering\small
\setlength{\tabcolsep}{5pt}
\begin{tabular}{@{}lrrr@{}}
\toprule
Model & Mean input tokens & Mean output tokens & Mean cost/task (\$)\\
\midrule
Clef-flash 9B & 181,277 & 0 & 0\\
Clef 27B & 276,130 & 0 & 0\\
GLiClass (0.4B) & 462,086 & 0 & 0\\
Jev & 264,195 & 14,503 & 0.0111\\
JevK5 4B & 933,259 & 0 & 0\\
Open-Jev-2B & 18,372,135 & 0 & 0\\
mini-Jev & 1,324,752 & 0 & 0\\
StartLux 4B & 634,567 & 0 & 0\\
StartLux 27B & 552,111 & 0 & 0\\
decision-model-preview & --- & --- & ---\\
MiMo v2.6 Ultra Speed & 81,899 & 365 & 0.3173\\
DeepSeek v4.1 Flash & 14,708 & 1,471 & 0.0321\\
GPT-6.1-sol & 224,842 & 6,117 & 0.5109\\
\bottomrule
\end{tabular}
\caption{Mean token usage and mean API cost per recorded task run. Zero output tokens are values reported by the decision interface; zero API fees exclude local compute costs. Dashes denote unavailable token usage or API billing. GLiClass, Clef, StartLux, JevK5, Open-Jev, and mini-Jev run locally and incur no API fees. JevK5, Open-Jev, and mini-Jev means use 80, 79, and 180 completed recorded runs, respectively. DeepSeek API cost includes recorded retries.}
\label{tab:api-cost}
\end{table*}
}

%% file: tables/benchmark-comparison.tex
\begin{table*}[!tb]
\centering
\footnotesize
\setlength{\tabcolsep}{4.5pt}
\resizebox{\textwidth}{!}{%
\begin{tabular}{@{}llllcccc@{}}
\toprule
Benchmark & & Suite & Clock & \shortstack{Live\\state} & \shortstack{Latency\\effect} & \shortstack{Paused\\control} & \shortstack{Pace\\sweep}\\
\midrule
SmartPlay & \citep{wu2024smartplay} & 6 games & Step & -- & -- & -- & --\\
BALROG & \citep{paglieri2025balrog} & 6 games & Step & -- & -- & -- & --\\
lmgame-Bench & \citep{hu2026lmgame} & 6 games & Step & -- & -- & -- & --\\
Learn2Play & \citep{li2026learn2play} & 20 games & Step & -- & -- & -- & --\\
GameWorld & \citep{ouyang2026gameworld} & 170 tasks & Paused & -- & -- & -- & --\\
\midrule
VideoGameBench & \citep{zhang2025videogamebench} & 10 games & Wall & $\checkmark$ & $\checkmark$ & $\circ$ & --\\
GameWorld-RT & \citep{ouyang2026gameworld} & 34 games & Wall & $\checkmark$ & $\circ$ & $\circ$ & --\\
RT Reasoning Gym & \citep{wen2026realtime} & 3 games & Token; wall & $\checkmark$ & $\checkmark$ & -- & $\circ$\\
\midrule
AgentBench & \citep{liu2024agentbench} & 8 domains & Serial & -- & -- & -- & --\\
WebArena & \citep{zhou2024webarena} & 812 tasks & Serial & -- & -- & -- & --\\
OSWorld & \citep{xie2024osworld} & 369 tasks & Serial & -- & -- & -- & --\\
OSWorld~2.0 & \citep{yuan2026osworld2} & 108 tasks & Wall (subset) & $\circ$ & $\circ$ & -- & --\\
Robotouille & \citep{gonzalez2025robotouille} & 30 templates & Step timers & -- & -- & -- & --\\
Gaia2 & \citep{froger2026gaia2} & 1,120 scenarios & Simulated & $\checkmark$ & $\checkmark$ & $\circ$ & --\\
\midrule
\rowcolor[gray]{0.95}
\textbf{RealtimeBench} & & \textbf{180 tasks} & \textbf{Wall} & $\checkmark$ & $\checkmark$ & $\checkmark$ & $\checkmark$\\
\bottomrule
\end{tabular}}
\caption{How benchmarks treat time. Step/Serial interaction advances after a response; Paused freezes inference; Wall uses wall-clock time; Token uses inference-token time; Simulated uses a virtual clock. \emph{Live state}: evolution during inference; \emph{latency effect}: the outcome cost of delay is measured. $\checkmark$: evaluated; $\circ$: partial (subset or related control); --: not evaluated.}
\label{tab:benchmark-comparison}
\end{table*}

%% file: figures/figure-mario-practice.tex
\begin{figure*}[!tb]
\centering
\includegraphics[width=\textwidth]{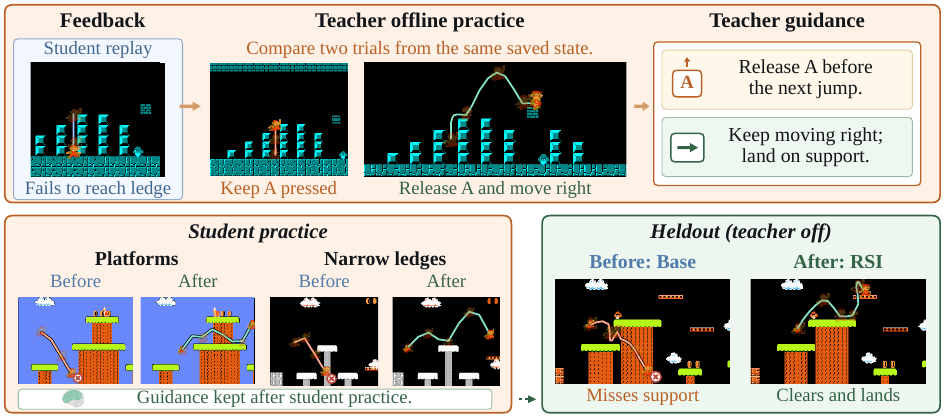}
\caption{The first Mario improvement round: feedback, teacher practice, draft-guidance checks by the student, and held-out outcomes. Base uses the initial guidance; R1 uses round-1 guidance. These panels illustrate the full guidance update, not the isolated effect of one rule.}
\label{fig:reasoning-guidance}
\end{figure*}

%% file: figures/teacher-reasoning-examples.tex
\begin{figure*}[!tb]
\centering
\includegraphics[width=\textwidth]{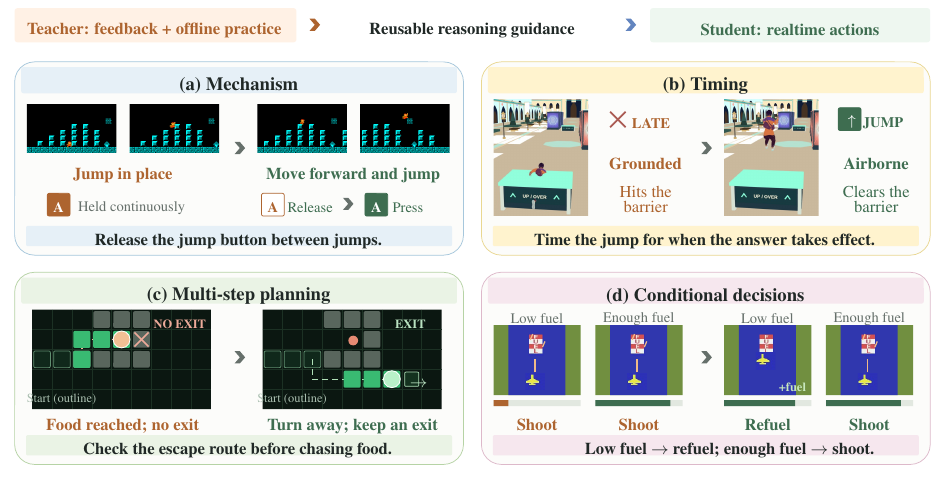}
\caption{Offline teacher practice turns reasoning experience into guidance for real-time student actions. In each card, the left side shows behavior before guidance and the right side shows behavior after guidance: (a) Mario releases jump input between jumps; (b) Temple times a jump to clear a barrier; (c) Snake preserves an escape route; (d) River conditions shooting or refueling on fuel level. Mario shows recorded paused branches from one feedback checkpoint; Temple, Snake, and River illustrate the rules rather than paired measured trajectories.}
\label{fig:guidance-types}
\end{figure*}

%% file: tables/clock-controls.tex
\begin{table*}[!tb]
\centering\small
\setlength{\tabcolsep}{7pt}
\resizebox{\textwidth}{!}{%
\begin{tabular}{@{}lrrrrrr@{}}
\toprule
 & \multicolumn{3}{c}{GPT-6.1-sol} & \multicolumn{3}{c}{Jev}\\
\cmidrule(lr){2-4}\cmidrule(lr){5-7}
Game type & Paused & 1.5$\times$ & 2$\times$ & Paused & 1.5$\times$ & 2$\times$\\
\midrule
Mario & 48.20 & 6.20 & 6.04 & 29.31 & 19.12 & 17.40\\
Snake & 42.71 & 0.12 & 0.12 & 39.65 & 2.47 & 1.65\\
Tetris & 57.08 & 0.00 & 0.00 & 26.65 & 19.92 & 4.87\\
Temple & 43.91 & 0.63 & 0.63 & 25.70 & 5.94 & 5.07\\
River & 38.00 & 0.03 & 0.00 & 41.59 & 10.15 & 6.17\\
Bird & 33.20 & 2.00 & 2.00 & 6.93 & 7.73 & 4.53\\
\bottomrule
\end{tabular}}
\caption{Test scores for the Paused and faster-clock conditions in Figure~\ref{fig:clock-gap}. GPT-6.1-sol uses medium reasoning effort. Each entry is the mean over ten tasks; where a condition was repeated, the entry is the mean over runs.}\label{tab:clock-native}
\end{table*}